# AutoPCF: Efficient Product Carbon Footprint Accounting with Large Language Models

Zhu Deng[1,#], Jinjie Liu[2,#], Biao Luo[1], Can Yuan[1], Qingrun Yang[1], Lei Xiao[1], Wenwen Zhou[1, *], Zhu Liu[3]

[1] Alibaba Cloud, Hangzhou, China

[2] School of Science and Engineering, The Chinese University of Hong Kong, Shenzhen, China

[3] Department of Earth System Science, Tsinghua University, Beijing, China

[#] These authors contribute equally.

[*] Corresponding: zhoubo.zww@alibaba-inc.com

**Abstract**

The product carbon footprint (PCF) is crucial for decarbonizing the supply chain, as it measures the direct and indirect greenhouse gas emissions caused by all activities during the product's life cycle. However, PCF accounting often requires expert knowledge and significant time to construct life cycle models. In this study, we test and compare the emergent ability of five large language models (LLMs) in modeling the 'cradle-to-gate' life cycles of products and generating the inventory data of inputs and outputs, revealing their limitations as a generalized PCF knowledge database. By utilizing LLMs, we propose an automatic AI-driven PCF accounting framework, called AutoPCF, which also applies deep learning algorithms to automatically match calculation parameters, and ultimately calculate the PCF. The results of estimating the carbon footprint for three case products using the AutoPCF framework demonstrate its potential in achieving automatic modeling and estimation of PCF with a large reduction in modeling time from days to minutes.

## 1   Introduction

Climate change and its adverse effects have become a pressing global concern, mandating immediate action for sustainable practices and mitigation of greenhouse gas emissions[1]. Product Carbon Footprint (PCF), which quantifies life cycle emissions associated with products, enables informed decision-making and the development of effective strategies for reducing environmental impact[2]. While Life Cycle Assessment (LCA) has been widely used to estimate the PCF, this approach often faces big challenges. For example, the determination of system boundaries and construction of complete inventories of products' life cycles necessitates extensive research, often relying on a combination of primary data collection, literature reviews, and expert judgment[3], which is time-consuming, resource-intensive, and subject to uncertainties[4]. Given the limitations of existing methods, exploring alternative approaches that enhance the efficiency and objectivity of carbon footprint estimation becomes necessary.

The emergence of large language models (LLMs), such as the GPT series[5–7] and GLM[8,9], provides the possibility of a rapid generation of professional knowledge. LLMs are powerful deep learning models that have been trained on vast amounts of text data, enabling them to generate coherent and contextually relevant responses. Moreover, they are often deployed on the cloud to improve response efficiency. When using pre-trained or fine-tuned LLMs for professional knowledge question answering, the models can leverage their inherent knowledge to enhance performance and efficiency in providing accurate and insightful responses. Therefore, LLMs have been widely applied in various domains, including software[10], finance[11], biology[12], medicine[13], geography[14], and carbon management[15–17]. The corpus used for training LLMs contains a wide range of texts, which may include supply chain information related to product carbon footprint (PCF), and have the potential to generate detailed descriptions and simulate various production processes[18–21]. This suggests the possibility of applying LLMs to PCF modeling and estimation to improve efficiency[22–24].

In this study, we propose an automatic AI-driven product carbon footprint accounting framework, called AutoPCF, which is primarily driven by LLMs. We employed five general LLMs to automatically generate the production processes and their use of raw materials and energy as well as their wastes, to establish 'cradle-to-gate' life cycle models. In this study, we applied AutoPCF to three products which are hot rolled round steel, printed and dyed fabric, and lithium iron phosphate battery. Combining with the semantics-based matching model, we matched the emission factors corresponding to raw materials and energy for each production process, and finally calculated their carbon footprint. The benchmark between the AutoPCF results and those estimated by our experts shows promise of using the AutoPCF framework to scale and accelerate the process of product carbon footprint accounting compared with traditional expertise-based processes.

## 2 Materials and Method

### 2.1 Scope definition

The accounting of PCF usually follows an LCA process, which generally includes five stages: raw material extraction, manufacturing, distribution, use, and end-of-life disposal. This accounting scope is also known as 'cradle-to-grave'[25]. In this study, the PCF accounting scope is 'cradle-to-gate', which mainly focuses on the production and manufacturing process of the product, as it is one of the most concentrated life cycles for raw materials and energy[26].

### 2.2 The traditional expertise-based process

Estimating PCF generally comprises three primary steps: determining system boundaries, conducting a life cycle inventory analysis, and evaluating environmental impacts. LCA analysts establish a life cycle inventory (LCI) based on their expert knowledge and experience, according to the defined system boundaries of the product's carbon footprint. Specifically, analysts typically dismantle the processes of each life cycle stage of the product and determine the inputs and outputs required for each process to ultimately build the product's life cycle model. They then evaluate the environmental impacts (i.e., greenhouse gas emissions) of all inputs and outputs throughout the product's life cycle. The input items of the product usually include raw materials and energy, and greenhouse gas emissions resulting from the production of raw materials, consumables, and secondary energy sources (e.g., electricity) that occur during the product's life cycle are included in the product's carbon footprint (Scope 2+Scope 3 upstream). Output items typically include waste gas, wastewater, and waste materials, such as greenhouse gas emissions resulting from the use of fossil fuels (Scope 1). Finally, these greenhouse gas emissions are converted into carbon dioxide equivalents ($CO_2$-eq) and added up to calculate the product's carbon footprint.

This process can be time-consuming and involves breaking down each life cycle stage of the product based on industry production experience or relevant literature to determine inputs and outputs for each process. Then, values for the input-output inventory are determined based on actual production conditions, and the results are ultimately calculated. Although there are some life cycle modeling tools and life cycle databases that can help improve calculation efficiency, it still takes several days or months to complete the carbon footprint assessment of a product.

## 2.3 AutoPCF: the AI-driven automatic modeling framework

The AutoPCF framework follows a similar process to the traditional expertise-based process of PCF modeling and estimation. However, unlike LCA, which requires expert knowledge for inventory construction and emission factor selection, AutoPCF leverages Artificial Intelligence (AI) and Deep Learning algorithms for these tasks. For inventory construction, AutoPCF primarily utilizes the capabilities of Large Language Models (LLMs). By providing appropriate prompts to LLMs, the framework generates and outputs all the production processes of a product. LLMs are also used to generate specific activity data for each production process. In terms of emission factor selection, AutoPCF employs semantic similarity to automatically match the activity name with the corresponding emission factors from a database, obtaining the emission factor value of each activity. The carbon footprint of the product is calculated based on the activity data generated by LLMs and matched emission factors. All of these steps can be automated through programming and encapsulation. By simply providing the name of a product, the framework can automatically generate the life cycle inventory and carbon footprint of the product.

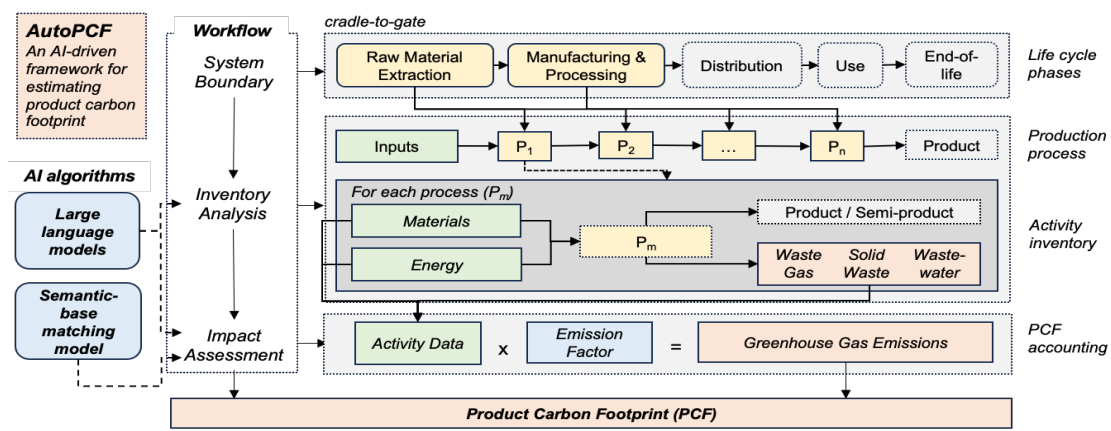

*Figure 1 | The AutoPCF framework.*

### 2.3.1 The 'cradle-to-gate' inventory analysis

*Introducing LLMs*

LLMs have shown excellent ability in a wide range of NLP tasks and can better obtain corresponding text content according to the research purpose by appropriately adjusting prompts and combining in-context learning[27,28]. During the training of LLMs, a significant amount of text corpus is used, such as GPT-4 with more than 1 trillion parameters. The GPT-3.5 and GPT-4 training datasets include the Common Crawl dataset, the Webtext2 dataset, two internet-based book corpora, and Wikipedia[7]. The Common Crawl dataset, which contributes 60% of the total weight in the training mix, contains petabytes of data obtained over 12 years of web crawling[29]. Plenty of data about carbon footprint can be found in it, e.g., data from carbonfootprint.com[30]. For GLM-130B, the training data include 1.2T Pile English corpus, 1.0T Chinese WudaoCorpora, and 250G web-crawled Chinese corpora[9]. Although specific descriptions of the training data for ChatGLM-6B and Tongyi Qianwen were not found, their datasets should also be adequately large to be prepared for carbon footprint assessment. Such a large amount of text corpus may equip the LLMs with knowledge of the production process and input-output activity inventory of products.

*Table 1 | Large language models used in this study.*

| LLM | Size | Description |
| --- | --- | --- |
| GPT-3.5[7,31] | 175B | Developed by OpenAI, GPT-3.5 is a set of models that improve on GPT-3. Gpt-3.5-turbo is the most capable GPT-3.5 model, with 4096 maximum tokens and training data up to Sep. 2021. |
| GPT-4[31,32] | >1T* | Developed by OpenAI, GPT-4 is more capable than any GPT-3.5 model, with 8192 maximum tokens and training data up to Sep. 2021. |
| GLM-130B[9] | 130B | Developed by Tsinghua University, GLM-130B is a bilingual language model. It outperforms GPT-3 175B in English and ERNIE TITAN 3.0 260B in Chinese. |
| ChatGLM-6B[33] | 6.2B | ChatGLM-6B is an open bilingual (Chinese and English) language model based on the GLM framework. It has some limitations due to its small size, including a lack of English ability. |
| Tongyi Qianwen[34] | >10B | Developed by Alibaba Cloud, Tongyi Qianwen is a large language model supporting both Chinese and English. It will be integrated into all business applications across Alibaba's ecosystem. |

* No official report. Estimated by users.

In this study, we employed 5 LLMs in AutoPCF, and evaluate their performance in the PCF modeling (See *Table 1*). The API for ChatGLM-6B is locally deployed in this study, while the APIs for the other four LLMs are all called through HTTP requests. We utilized the generation ability of LLMs to quickly perform product lifecycle inventory analysis by guiding LLMs to break down the production process and activity inventory of the product through appropriate prompts. This breakdown process is conducted in a conversational way by gradually providing prompts and incorporating contextual information.

*Inventory analysis*

Using LLMs for inventory analysis involves two steps: production process breakdown and activity inventory construction. Firstly, we perform a production process breakdown for the product. We first construct a prompt that predefines the question and output requirements, providing a template for the generation of subsequent results (see **SI Text 1**). Next, we make full use of the contextual learning ability of LLMs to replace the product name in the template, and LLMs output the production process of the product. Then, for each production process, we let LLMs generate the activity inventory (i.e., a list of raw material inputs, energy consumption, and waste generated). Similarly, we first construct a prompt containing input-output templates, and then let LLMs output the activity inventory list for that production process (see **SI Text 1**). The prompts used for various LLMs are generally kept consistent and are not specifically optimized for individual models.

To evaluate the performance of the inventory analysis by LLMs compared to the expert-based process, we follow the ROUGE method[35] and define three metrics, $recall$, $precision$, and $F1\text{-}Score$. The $Precision$ is defined as the fraction of correct retrieved instances among all the retrieved instances (instance here means the item in the inventory, such as coke breeze, pig iron, oxygen, etc.). The $Recall$ is the ratio between the number of the correct retrieved instances and all correct instances. The $F1\text{-}score$ is the harmonic mean of precision and recall, which can measure the overall performance of the model.

$$precision = \frac{the\ number\ of\ correct\ retrieved\ instances}{the\ number\ of\ all\ retrieved\ instances} \quad (1)$$

$$recall = \frac{the\ number\ of\ correct\ retrieved\ instances}{the\ number\ of\ all\ correct\ instances} \quad (2)$$

$$F1\text{-}score = 2 * \frac{precision*recall}{precision+recall} \quad (3)$$

*Generating activity data of the inventory*

Activity data refers to the amount of raw materials or energy consumed during the production process. In traditional LCA assessment, activity data is collected through automatic monitoring or questionnaires. In this study, we used two methods to generate activity data required for PCF accounting, including the direct generation approach and the indirect estimation approach.

Direct generation approach: We constructed appropriate prompts to enable LLMs to automatically generate and output activity data corresponding to the inventory (see **SI Text 1**). However, not all LLMs are capable of generating seemingly reasonable activity data. Our results suggest using GPT-3.5 to directly generate activity data.

Indirect estimation approach: It employed a top-down method based on industry-wide macro-statistics data to generate activity data for each process in the emission inventory. We matched the activities in the inventory with the industries in the input-output table to convert the input economic quantity into a physical quantity, which determines the amount of raw materials or energy input (see **SI Text 4**). This approach was applied to all LLMs for activity data estimation. However, there is in fact a large discrepancy between the actual input-output of product production and the industry-wide input-output, resulting in a considerable level of uncertainty in the calculated results when using this approach.

### 2.3.2 Product carbon footprint assessment

**Semantic-based emission factor matching model**

To estimate the environmental impact of each activity in the inventory, it is generally necessary to multiply the activity data by the emission factor (i.e., the amount of greenhouse gas emissions per unit of activity data) to convert the activity data into greenhouse gas emissions. In this study, we employed a semantic-based model to match the aforementioned activities with their corresponding emission factors. The emission factor database used in this study is the Ecoinvent v3.9.1 database[36]. The names of both the emission factors and activities are first encoded into vectors with a length of 768 by a pre-trained language model. We calculated the semantic similarity between each activity vector and all emission factor vectors (**Eq**. (4)), identifying the emission factor with the shortest vector distance (considered as the highest similarity) to match it with the corresponding activity as its emission factor.

$$\cos\_sim(\vec{a},\vec{f}) = \frac{\vec{a}\cdot\vec{f}}{|\vec{a}|\cdot|\vec{f}|} \qquad (4)$$

where $\cos\_sim(\vec{a},\vec{f})$ is the cosine similarity between the activity vector and emission factor vector, $\vec{a}$ and $\vec{f}$ are the activity vector and emission factor vector.

### 2.4 Product carbon footprint accounting

In PCF accounting, the first step is to break down the production process of the product life cycle, and obtain the inputs, energy use, and outputs of each process. The greenhouse gas (GHG) emissions generated by each process can be calculated using the following formula:

$$E_s = AD_f * EF_f \qquad (5)$$

Where $E_s$ denotes the GHG emissions from process $s$; $AD_f$ and $EF_f$ are the activity data and the corresponding GHG emission factor of the activity $f$, such as the raw materials input or energy consumption in the process[37].

The PCF assessment takes into account seven greenhouse gases, namely carbon dioxide ($CO_2$), methane ($CH_4$), nitrous oxide ($N_2O$), hydrofluorocarbons (HFCs), perfluorocarbons (PFCs), sulfur hexafluoride ($SF_6$), and nitrogen trifluoride ($NF_3$), which are covered by the UNFCCC/Kyoto Protocol. To provide a standardized measure, the PCF is expressed in terms of $CO_2$ equivalent ($CO_2$-eq) based on their respective Global Warming Potentials (GWPs).

The estimation by AutoPCF models is measured by the error (**Eq. 6**) and adjusted coefficient of variation (adjusted CV, **Eq. 7**) compared to the expert-based results.

$$error = \left|1 - \frac{PCF_{AutoPCF}}{PCF_{expert-based}}\right| \qquad (6)$$

$$adjusted\ CV = \frac{\sigma}{\mu} = \frac{\sqrt{\frac{\sum_{i=1}^{n}(x_i-\bar{x})^2}{n-1}}}{\frac{\sum_{i=1}^{n}x_i}{n}}, Q1 - 1.5 * IQR \leq x_i \leq Q3 + 1.5 * IQR \tag{7}$$

### 2.5 The three case products

We selected three products as the main case studies for our research. We previously used the Energy Expert platform (https://energy.alibabacloud.com), a product carbon footprint modeling tool, to conduct a life cycle assessment and carbon footprint calculation for these three products. Therefore, we have a good understanding of the life cycle processes of these three products, which allows us to better evaluate the production processes and inventory analysis of AutoPCF. Additionally, we reviewed other literature on the carbon footprint estimation of these three products (hot rolled round steel[38], printed and dyed fabric[39,40], and lithium iron phosphate battery[41–46]) and determined the estimated median and uncertainty range based on expert-based models to evaluate the error and uncertainty of the AutoPCF results.

## 3 Results

### 3.1 The emergent ability of LLMs on the 'cradle-to-gate' product inventory analysis

As shown in **Fig 2**, we use precision (P), recall (R), and F1-Score (F1) as evaluation metrics to assess the differences between the LLMs generated results and the expert-based process (see ***Materials and Method***). The results showed that for the production process, GPT-3.5 and GPT-4 had a higher average F1-score of 0.59 and 0.53 for the three products than other models, indicating their better performance in modeling product production processes. In comparison, the average F1-score of GLM-130B and ChatGLM-6B were only 0.20 and 0.37, respectively. Specifically, the precision of GPT-4 results was the highest, reaching 56%, indicating that the GPT-4 decomposition results are closer to the expert-based production process. The recall of GPT-3.5 was high (72%) but the precision was a little lower (52%), indicating that although GPT-3.5 found as many production processes as possible, many of the generated answers are not within the expert-based process. In comparison, the average precision and recall are only 41% and 18% for GLM-130B and 49% and 30% for ChatGLM-6B respectively.

For the activity inventory, GPT-3.5 and GPT-4 also outperformed the other models in all three metrics, with a precision exceeding 59%, recall exceeding 32%, and F1-Score above 0.4 (GPT-3.5 achieved an F1-Score of 0.53 for modeling the inventory of printed dyed fabric, and GPT-4 reached an F1-Score of 0.50). This indicates that GPT-3.5 and GPT-4 perform relatively better in modeling the activity inventory of these three products. In comparison, GLM-130B and GLM-6B showed relatively poorer performance in all three metrics, especially for hot rolled round steel, where F1-Score was only 0.13 and 0.1, respectively.

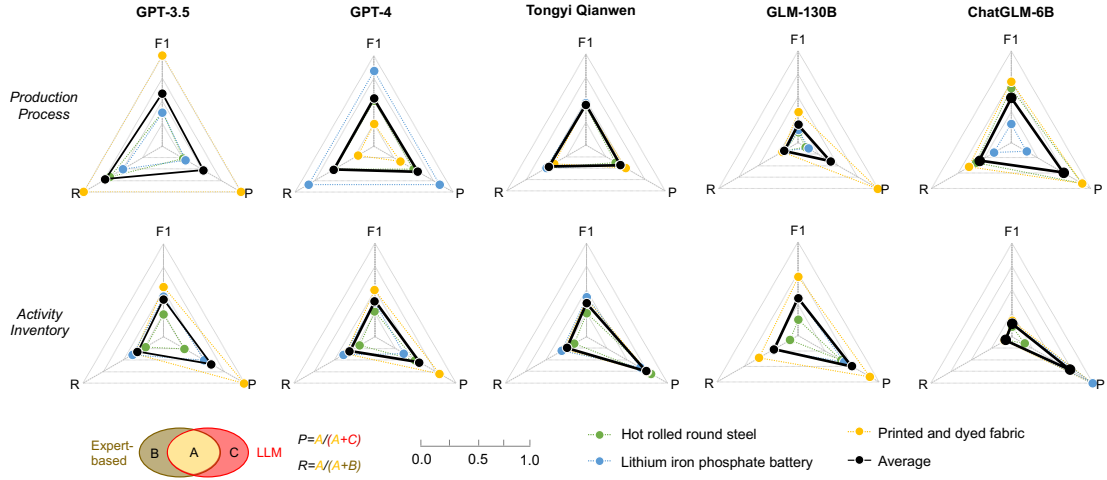

*Figure 2 | Performance of inventory analysis by using the five large language models (i.e., GPT-3.5, GPT-4, Tongyi, GLM-130B, and GLM-6B). The performance is evaluated by three metrics, precision (P), recall (R), and F1-score (F1), all ranging from 0 to 1, with 1 being the best and 0 being the worst compared to the expert-based process.*

### 3.2 The AutoPCF estimations of the three product cases

Based on the production processes and activity inventory generated by the five LLMs, we accordingly generate activity data (amount of input and output of each process) by two approaches: the direct generation approach (DGA) and the indirect estimation approach (IEA) (see ***Materials and method***). The carbon footprint of the three products is then estimated based on the generated activity data. For each of the ten models (five LLMs using two activity data generation methods), we repeated the estimation process 20 times to assess the stability of the prediction results, and their medians and ranges were compared with the result of the expert-based model (**Fig 3**).

In general, the AutoPCF_GPT-3.5_DGA model demonstrated PCF estimation that is closer to the median of expert-based results. For hot rolled round steel, the AutoPCF_GPT-3.5_DGA model showed the smallest error among all models within 10%, with a median estimation value of 2.1 kgCO2-eq/kg compared to the median estimation of 2.3 kgCO2-eq/kg by expert-based models. However, for printed and dyed fabric (PDF) and lithium iron phosphate battery (LIPB), the errors were larger (35% for PDF and 85% for LIPB, respectively). Nonetheless, the AutoPCF_GPT-3.5_DGA model still exhibited the smallest average error (42%) of these three products among all models whether the activity data generated by DGA (AutoPCF_GPT-4_DGA: 117%, AutoPCF_Tongyi_DGA: 344%, AutoPCF_GLM-130B_DGA: 110%, AutoPCF_GLM-6B_DGA: 57%) nor IEA (AutoPCF_GPT-3.5_IEA: 98%, AutoPCF_GPT-4_IEA: 54%, AutoPCF_Tongyi_IEA: 497%, AutoPCF_GLM-130B_IEA: 69%, AutoPCF_GLM-6B_IEA: 80%).

Although some of the AutoPCF models output results that appear closer to the expert-based model results but exhibit poorer stability. For instance, the AutoPCF_GLM-130B_DGA model has the highest instability among all models with the adjusted coefficients of variation (adjusted CV) of 149% for HRRS, 84% for PDF, and 53% for LIPB. This instability may be attributed to the process of generating activity data. The adjusted CV for the IEA-based model (AutoPCF_GLM-130B_IEA) was only 110% for HRRS, 57% for PDF, and 49% for LIPB, suggesting that the pre-trained GLM-130B model may lack prior knowledge of activity data, leading to inaccurate estimations and unstable results.

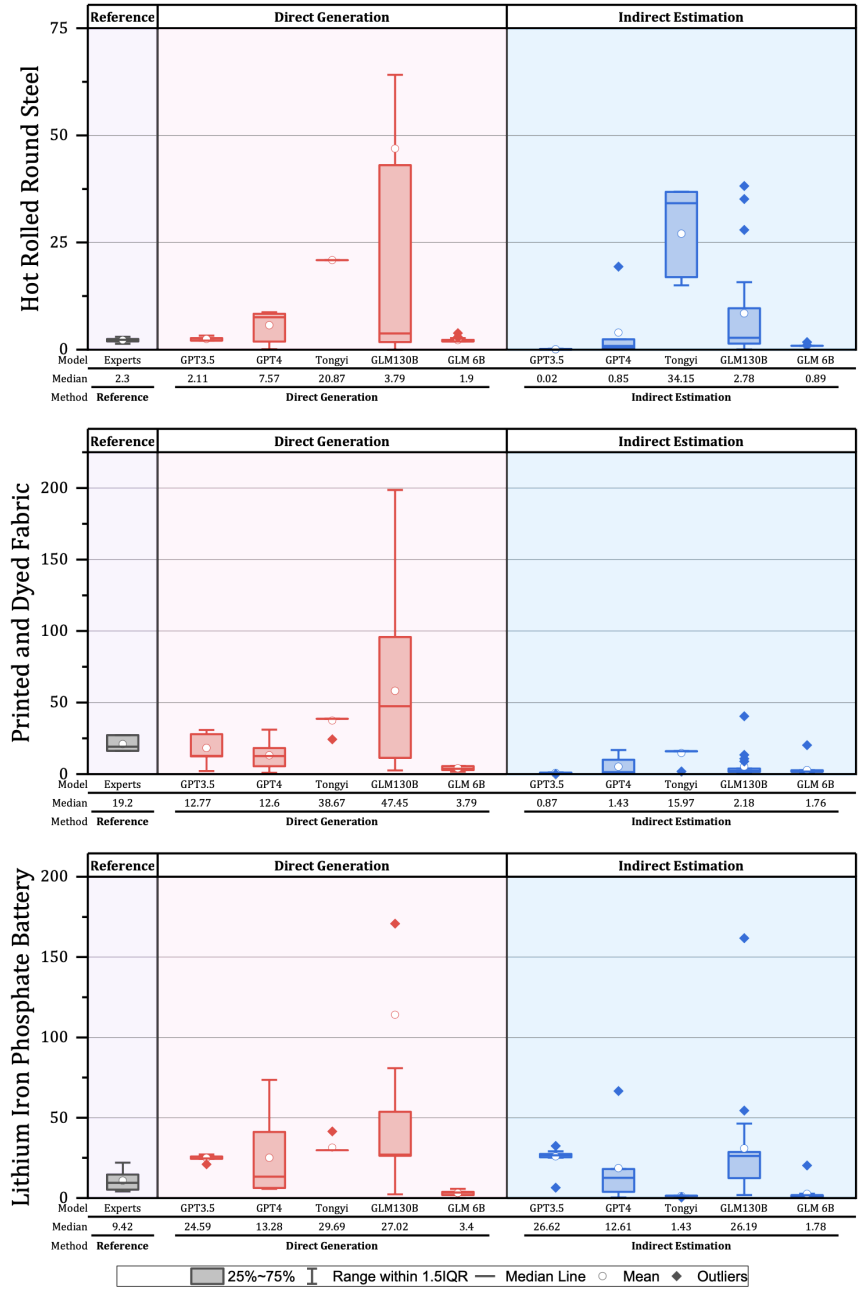

*Figure 3 | Carbon footprint estimation of the three selected products (hot rolled round steel, printed and dyed fabric, and lithium iron phosphate battery).* The gray bars denote the 25% to 75% carbon footprint estimation range estimated by the expert-based models, while the activity

*data is collected from industrial expertise ('Reference'). The red bars denote the 25% to 75% carbon footprint estimation range of hot rolled round steel, printed and dyed fabric, and lithium iron phosphate battery, with both the inventory analysis and the activity data directly generated by LLMs. The blue bars denote the 25% to 75% carbon footprint estimation range of hot rolled round steel, printed and dyed fabric, and lithium iron phosphate battery, with the inventory analysis conducted by the LLMs and using the activity data estimated by industrial proxy data. The error bar represents the 1.5 interquartile range (IQR) of the estimation results for different models.*

### 3.3 The generalization potential of the AutoPCF

After analyzing the performance of ten models, the AutoPCF_GPT-3.5_DGA model exhibited relatively good stability and estimation for the three products. Therefore, we selected 20 industrial products and used the AutoPCF with GPT-3.5 model to estimate their PCFs. The results showed that the AutoPCF_GPT-3.5_DGA model potentially outputs a larger prediction of carbon footprint among these products compared to the expert-based database[47] (**Fig 4**). Among these products, AutoPCF-GPT-3.5 demonstrated better performance for certain products, including section steel, nonwoven fabric, PVC, and automotive aluminum alloy, with an error within 20% compared to expert-based estimations. In particular, the PCF estimation discrepancy for section steel by AutoPCF_GPT-3.5_DGA was significantly low compared to human experts. For common products with relatively uniform production processes, LLMs can provide accurate responses based on information extracted from sufficient relevant training data. However, AutoPCF might result in high carbon footprint estimation errors for certain products, such as alumina. We found that the substantial error in LLMs' inventory and activity data for alumina production was due to higher electricity consumption values for each production process, leading to PCF overestimation. This discrepancy highlights the need to investigate methods that enhance the generalization capabilities of AutoPCF. While LLMs are powerful language models capable of generating human-like text, they may provide incorrect information when faced with unfamiliar prompts. Therefore, considering the model's limitations and the quality of its training data is crucial to ensure reliable and accurate results when utilizing LLMs for specific tasks, such as estimating PCFs.

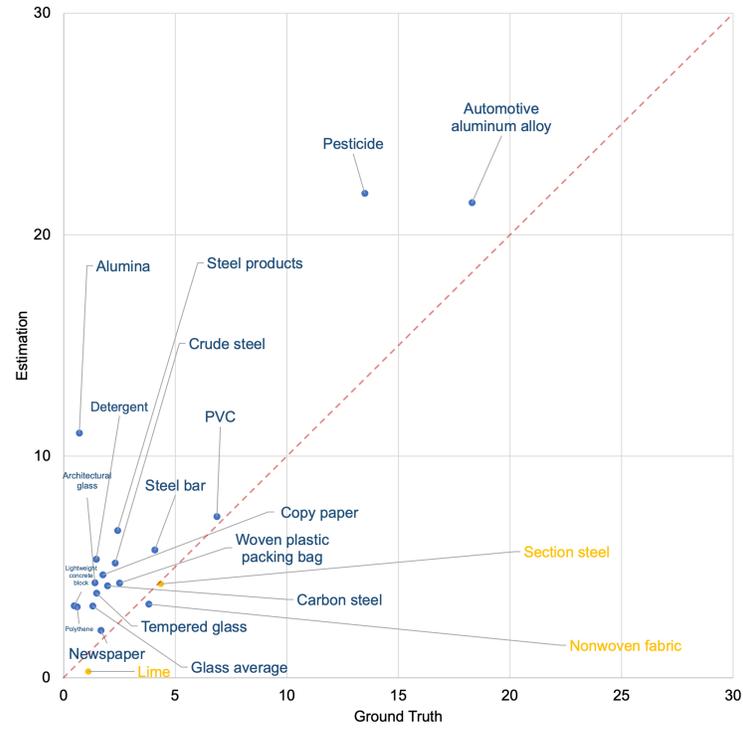

*Figure 4 | Carbon footprint estimation of 20 industrial products based on the AutoPCF_GPT-3.5_DGA model (detergent, lightweight concrete block, lime, newspaper, crude steel, section steel, copy paper, steel bar, woven plastic packing bag, potash fertilizer, alumina, tempered glass, architectural glass, carbon steel, polythene, nonwoven fabric, PVC, pesticide, steel products, automotive aluminum alloy, glass average).* The inventory analysis and activity data are directly generated by GPT-3.5, and the ground truth value is given by the CPCD database[47]. The dotted red line represents the perfect estimation with no error compared to the ground truth. The yellow dots denote the carbon footprint of three products (section steel, lime, and nonwoven fabric), with the AutoPCF_GPT-3.5_DGA model probably underestimating their carbon footprints compared to the ground truth. The blue dots denote the carbon footprint of the other 18 products, with a potential overestimation.

## 4 Discussion

### 4.1 Uncertainty of the AutoPCF estimations

The uncertainties in expert-based PCF estimation stem from diverse factors, including the intricate and variable nature of production processes, data availability and quality, and subjective judgment when making assumptions. Regarding the emission inventory, experts were tasked with compiling a comprehensive and detailed list of GHG emissions associated with various sources throughout the product's life cycle. The accuracy and completeness of this emission inventory can vary based on available data and the level of expertise of the experts. Subjective judgment in the estimation process can introduce variations and uncertainties in the final results. Uncertainties may arise from incomplete or missing data on emission sources. The reliability of activity data, such as energy consumption, material inputs, and process parameters, played a crucial role in PCF estimation. Experts often rely on available data, historical records, or industry standards to estimate these parameters. However, the need to use proxy data or generalize information can introduce uncertainties due to variations in data quality and the absence of specific data for certain processes. Estimating emission factors, which quantify GHG emissions per unit of activity or material, can also be challenging. Experts might draw upon emission factors from different sources or adapt them to specific contexts, leading to uncertainties in PCF estimation. Moreover, variations in the production of products across different factories, regions, or countries further compound the uncertainty. Experts may need to make assumptions or use models to estimate emissions for certain stages, and the selection of these assumptions and models can introduce uncertainties in the final PCF estimation.

Due to the stochastic nature of LLMs, which can provide different answers for the same question under multiple inquiries, ensuring stability became challenging and introduced performance inconsistencies. Nevertheless, when estimating activity data directly or indirectly, AutoPCF based on GPT-3.5, ChatGLM-6B, or Tongyi Qianwen demonstrated lower uncertainty compared to human experts for the selected three products, indicating their reliable stabilities. However, AutoPCF with GLM-130B or GPT-4 tended to introduce larger uncertainties, particularly in the PCF estimation of printed and dyed fabric and lithium iron phosphate battery. The use of an industry matching model with an accuracy of only 74% for indirect estimation contributed to the observed large uncertainty. The results revealed that AutoPCF based on ChatGLM-6B, Tongyi Qianwen, or GPT-3.5 with indirect activity data generation exhibited lower uncertainty compared to that with direct activity data generation, underscoring the importance of accurate activity data collection. Moreover, since the semantic-based emission factor model achieves an accuracy of 92%, the mismatched emission factors for some emission sources can also contribute to uncertainties. Despite these potential sources of uncertainty, AutoPCF based on GPT-3.5 and direct activity data generation consistently provided a PCF estimation range comparable to that of human experts.

In addition to the aforementioned issues, the conversion of units may also lead to potential errors in PCF estimation. For example, PCF assessment for lithium iron phosphate batteries is commonly expressed in kgCO2-eq/kWh. However, for the purpose of harmonization, we have transformed this measure into kgCO2-eq/kg by using the relationship between battery capacity and weight, aligning it with battery weight. This conversion introduces potential errors due to the diverse array of battery materials and weights. Further investigation into the intricacies of production processes and emission inventories reveals that this divergence was also influenced by variations in production methods and emission sources. AutoPCF's outcomes often assumed a production scenario involving the recycling of used batteries, wherein the high emission factors of these recycled batteries contributed to the higher PCF estimation results. Therefore, it is essential to carefully consider the conversion of units and thoroughly examine the production scenarios to ensure accurate and reliable PCF estimation results.

### 4.2 Improved efficiency by AutoPCF

The emission inventory comprises a comprehensive and detailed list of GHG emissions linked to various emission sources. The consistency and completeness of the emission sources can reflect the accuracy of the LLM model in identifying emission inventories, and further significantly influence the results of PCF estimation *(see **Fig 2 and Fig 3**)*. GPT-3.5 and GPT-4 exhibited better performance in generating both emission inventory and activity data, leading to more accurate PCF results when the AutoPCF is based on these two LLMs.

The proposed AutoPCF not only provides considerably believable estimation results but also demonstrates remarkable cost and time efficiency compared to human experts. Traditionally, experts relied on their experience from previous studies to compile the emission inventory and estimate PCF, resulting in variations and potential discrepancies. The intricate nature of production processes, facilities, materials, and products adds complexity and time consumption to precisely determine the exact emission sources. In contrast, AutoPCF offers an automatic and streamlined approach for PCF estimation, significantly reducing the need for extensive manual labor and saving valuable time and costs. Leveraging the power of LLMs, AutoPCF can tap into vast amounts of text data to learn from the collective knowledge, while deep-learning models rapidly output accurate information. These factors together contribute to the high efficiency of the AutoPCF framework, making it a promising and valuable method for estimating PCF and driving advancements in environmental research. For instance, the quick PCF estimation capabilities enabled by AutoPCF allow for continuous tracking of corporate carbon footprints and large-scale promotion. This opens up new possibilities for dynamic carbon footprint management, enabling timely interventions and targeted carbon reduction strategies.

However, the observed disparities in the emission inventory and PCF results still raise questions regarding whether our proposed AutoPCF and LLMs can be considered complete substitutes for human expertise, warranting further investigation into this matter. Future work can focus on enhancing the performance and accuracy of the underlying deep-learning models and LLMs used in AutoPCF. This may involve exploring advanced architectures, incorporating more diverse training data, and fine-tuning the models to specific industry sectors. Especially, the LLM models can be improved by constructing a local knowledge base and training an LLM model specially designed for PCF estimation. This integration of external knowledge can enhance the accuracy of the generated answers by reducing the inclusion of irrelevant or incorrect information. Note that ensuring the high quality and suitability of the knowledge base also plays a crucial role in achieving more accurate and reliable results. Optimizing and customizing prompts can help better extract information and improve the accuracy of emission inventory. Moreover, a novel PCF estimation model completely dependent on LLM models is to be studied by designing reasonable prompts, like AutoGIS[14]. Improving the interpretability of the models can also enhance stakeholders' trust in the generated carbon footprint estimates.

**4.3 Generalization**

The generalization capacities of AutoPCF and LLMs in estimating PCF can be both promising and challenging. LLMs, with their ability to learn from vast amounts of text data, can capture complex patterns and relationships in language, making them potentially adept at understanding and estimating PCF-related information from diverse sources. However, LLMs may lack specialized domain knowledge specific to PCF estimation, especially when dealing with complex industrial processes or specific product categories. The generalization capacity of LLMs depends on the availability and diversity of data used during their pre-training. Limited access to high-quality PCF-related data may affect their accuracy and generalization to new scenarios and different products. In addition, LLMs may inadvertently produce biased estimations in PCF if the training data contain biases, leading to inaccurate and unreliable results. While LLMs can generalize well within their pre-training domains, transferring their knowledge to specific PCF estimation tasks may require fine-tuning or adaptation to task-specific data.

Reiterate that PCF needs to be calculated based on real product production processes and activity data from the corresponding factories. It is important to acknowledge that even human experts cannot achieve 100% consistency of the inventory and PCF results due to the complexity and variability of production processes and raw materials. The objective here is to obtain the most comprehensive estimate of the carbon footprint of a product. Based on the estimated PCF, the standardized PCF default values are to be determined and can be used as a reference for environmental disclosure, especially for a producer who cannot provide credible and transparent PCF calculations.

It is important to acknowledge that even human experts cannot achieve 100% consistency of the inventory and PCF results due to the complexity and variability of production processes and raw materials. Variations in inventory construction, activity data, and emission factor selection can lead to differences in the estimated results of product carbon footprints, as evidenced by the varying results of expert models (*Fig 3*, the adjusted CV: 24% for HRRS, 27% for PDF, and 58% for LIPB). The objective here is to obtain the most comprehensive estimate of the carbon footprint of a product. Based on the estimated PCF, the standardized PCF default values are to be determined and can be used as a reference for environmental disclosure, especially for a producer who cannot provide credible and transparent PCF calculations.

## 5 Conclusion

In this study, we proposed an AI-driven automatic modeling framework, the AutoPCF, which utilizes large language models (LLMs) to assist LCA practitioners in estimating product carbon footprints (PCF) with enhanced efficiency and reduced costs. We built and tested a set of ten models using five different LLMs and two different activity data generation approaches to estimate the PCFs of three typical industrial products under the AutoPCF framework. The results were compared to expert-based estimations. Our findings revealed that some pre-trained LLMs, such as GPT-3.5, GPT-4, and Tongyi Qianwen, demonstrated good knowledge of production processes and activity lists, with relatively better performance in modeling production processes and activity inventories compared to the expert-based model. Specifically, the F1-score of over 0.4 for modeling production processes and over 0.35 for modeling activity inventories indicated the emergent potential of these three LLMs in product life cycle modeling and activity data generation. Among all models, the AutoPCF_GPT-3.5_DGA model exhibited relatively good stability and estimation for all three products, with the smallest average error of 42%. These results highlight the potential of LLMs in product life cycle modeling and activity data generation, while also emphasizing the need to enhance their generalization capabilities. When utilizing LLMs for specific tasks, such as estimating PCFs, it is crucial to consider their limitations and the quality of their training data to ensure reliable and accurate results. However, the automatic PCF modeling approach of AutoPCF allows for the rapid estimation of the carbon footprint of a specific product category with minimal data input, making it a versatile and efficient tool for promoting sustainability practices and supporting decision-making towards global environmental goals.

Future work aims to reduce the uncertainty and enhance the reliability of the AutoPCF in estimating PCF. Several key areas can be addressed to achieve these goals. Firstly, enhancing the training data for the LLMs used in AutoPCF is essential by incorporating a broader and more diverse range of texts related to PCF estimation. This will enable the models to better comprehend and interpret specific prompts, resulting in more accurate responses. Secondly, fine-tuning LLMs on domain-specific data related to carbon footprint estimation can enable the models to adapt and specialize for specific tasks, thereby improving performance on relevant prompts. Thirdly, crafting prompts carefully to provide clear and specific instructions to the language models should be explored. Fine-tuning prompts based on the particular model's strengths and weaknesses can yield more accurate and reliable results. Additionally, improving the accuracy of the industry matching model and emission factor matching model is necessary to reduce uncertainty stemming from activity data collection and emission factor matching. Fourthly, efforts should focus on improving the quality and completeness of the emission inventory data used in the estimation process. Relying on reliable and up-to-date emission factors and data sources will help reduce uncertainties in the emission inventory. Moreover, collaboration among experts, peer review, and transparent documentation of methods and assumptions can significantly contribute to uncertainty reduction and enhance the accuracy of PCF estimation. Involving domain experts in the field of PCF estimation to provide insights, validate the results, and address uncertainties will foster a collaborative effort that leads to more accurate estimations.

**Declaration of interests**

The authors declare no competing interests.

26. Niccolucci, V., Rugani, B., Botto, S., and Gaggi, C. (2010). An integrated *footprint* based approach for environmental labelling of products: the case of drinking bottled water. Int. J. DNE *5*, 68–75. 10.2495/DNE-V5-N1-68-75.

27. Lu, Q., Qiu, B., Ding, L., Xie, L., and Tao, D. (2023). Error Analysis Prompting Enables Human-Like Translation Evaluation in Large Language Models: A Case Study on ChatGPT. 10.48550/arXiv.2303.13809.

28. Cheng, L., Li, X., and Bing, L. (2023). Is GPT-4 a Good Data Analyst? 10.48550/arXiv.2305.15038.

29. Want to use our data? – Common Crawl https://commoncrawl.org/the-data/.

30. Common Crawl Index http://index.commoncrawl.org/CC-MAIN-2021-39-index?url=carbonfootprint.com/*.

31. OpenAI Platform https://platform.openai.com.

32. OpenAI (2023). GPT-4 Technical Report. 10.48550/arXiv.2303.08774.

33. ChatGLM-6B (2023).

34. Tongyi Qianwen https://qianwen.aliyun.com/.

35. Lin, C.-Y. (2004). ROUGE: A Package for Automatic Evaluation of Summaries.

36. Wernet, G., Bauer, C., Steubing, B., Reinhard, J., Moreno-Ruiz, E., and Weidema, B. (2016). The ecoinvent database version 3 (part I): overview and methodology. Int J Life Cycle Assess *21*, 1218–1230. 10.1007/s11367-016-1087-8.

37. Shi, X., and Meier, H. (2012). Carbon Emission Assessment to Support Planning and Operation of Low-carbon Production Systems. Procedia CIRP *3*, 329–334. 10.1016/j.procir.2012.07.057.

38. Khalid, Y., Wu, M., Silaen, A., Martinez, F., Okosun, T., Worl, B., Low, J., Zhou, C., Johnson, K., and White, D. (2021). Oxygen enrichment combustion to reduce fossil energy consumption and emissions in hot rolling steel production. Journal of Cleaner Production *320*, 128714. 10.1016/j.jclepro.2021.128714.

39. Fishwick, M. (2012). A Carbon Footprint for UK Clothing and Opportunities for Savings.

40. Institute for Prospective Technological Studies (Joint Research Centre), Beton, A., Cordella, M., Perwueltz, A., Desaxce, M., Gibon, T., Dias, D., Boufateh, I., Farrant, L., Kougoulis, J., et al. (2014). Environmental improvement potential of textiles (IMPRO Textiles) (Publications Office of the European Union).

41. Tao, Y., Sun, T., and Wang, Z. (2023). Uncovering various paths for environmentally recycling lithium iron phosphate batteries through life cycle assessment. Journal of Cleaner Production *393*, 136263. 10.1016/j.jclepro.2023.136263.
18

# Supplementary Information

**SI Text 1 | Inventory analysis by using large language models**

Large language models have shown excellent ability in a wide range of NLP tasks and can better obtain corresponding text content according to the research purpose by appropriately adjusting prompts and combining in-context learning. The characteristics of LLM can help researchers quickly determine the system boundary and emission inventory of target products in the process of LCA carbon footprint calculation.

The steps for constructing the production process and emission inventory are as follows:

1) **Construct the production process of a product**

```
prompt_to_generate_process=PromptTemplate(
    template="""
    You are now an expert in the production of "{product_name}".
    Please tell me the process stages involved in the "{product_name}" production process and return the results as follows:
    1. Output the result in JSON format.
    2. The result is a list containing only the names of the production process, such as ["process A", "process B", "process C"]
    3. If you can't get an answer from known information, return "None".
    """,
    input_variables=["product_name"]
)
```

A) First, we defined LLM as an expert in the field of production of our target product.

B) Second, we guided LLM to answer the process stage in the production of the target product.

C) And then, in order to facilitate subsequent data processing, we specify a structured output format like a JSON list.

D) Besides, in order to avoid the generation of false information as much as possible, we asked it to return "None" when it did not have enough knowledgeable information.

2) **Construct the life cycle inventory of a product**

```
prompt_to_generate_inventory = PromptTemplate(
    template="""
    You are now an expert in the production of "{product_name}". The production process of "{product_name}" can be divided into the following stages: {process_list}. Based on your knowledge, please tell me what raw materials and energy are needed, what products, wastewater, solid waste, and waste gas are produced in the "{current_process}" process of producing 1{unit} of "{product_name}", and provide the corresponding quantities of raw
```

> *materials, energy, products, wastewater, solid waste, and waste gas in JSON format. Please note the following requirements:*
>
> *1. The entity types included in JSON must belong to: [Product, Raw material, Energy, Waste gas, Wastewater, Solid waste].*
>
> *2. The JSON format conforms to the following form: {output_format},*
>
> *3. The units used in the results are all in international standard units, belonging to [kg, kWh, m, m3, MJ].*
>
> *4. If the substance in "Raw material" comes from the previous process, it should be noted which process it comes from in "source"; if not, "source" should be "None".*
>
> """,
>     input_variables=["product_name", "process_list", "current_process", "unit", "output_format"]
> )
>
> output_format = {"Product": [{"name": "A", "quantity": 1, "unit": "kg"}],
>                  "Raw material": [{"name": "B", "quantity": 1, "unit": "kg", "source": "process"}],
>                  "Energy": [{"name": "C", "quantity": 1, "unit": "kWh"}],
>                  "Wastewater": [{"name": "D", "quantity": 1, "unit": "kg"}],
>                  "Solid waste": [{"name": "E", "quantity": 1, "unit": "kg"}],
>                  "Waste gas": [{"name": "F", "quantity": 1, "unit": "kg"}]}

A) In this part, we defined LLM as an expert in the field of production of our target product as well.

B) And then, we asked LLM to answer the detailed information of each process in the answer list of step 1). The detailed information included the input (raw material and energy) and output (product, wastewater, solid waste and waste gas) inventories.

C) For processing data easier, we specify a structured output format like a JSON dictionary, and also rule the unit in the result.

D) Considering that the raw materials of part of the process may come from the previous process, we use the filed "source" for marking.

**SI Text 2 | Indirect estimation approach of activity data by using industrial proxy data**

1) Industrial classification matching

To address the issue of inconsistent industrial classification categories across different assessment processes, an approach is adopted to harmonize and align the categories. In this paper, the breakdown of product carbon footprints (PCFs) is initially presented using the Central Product Classification (CPC) categories. However, Chinese macro-statistics data typically follow the industrial classification system for national economic activities, known as GB/T 4754.

To bridge this gap and determine the activity data for the corresponding industry, a deep-learning based semantic matching model is employed. This model enables the alignment of other classification categories with the CPC framework. By mapping the emission inventory's CPC categories to the relevant activity data category, the corresponding industry-specific activity data is identified. This ensures consistency and accuracy in linking the emission inventory with the appropriate activity data, enabling a comprehensive assessment of the PCF for different industries.

2) Activity data of raw material

Input-Output Tables (IOTs) describe the sale and purchase relationships among industries. For any certain industry, IOTs provide the value of raw materials per unit of product acquired from other industries. The amount of raw material is calculated based on the unit value of products, such that

$$m_{ij} = \frac{v_{ij}v_i}{v_j} = \sum_k m_{ijk}$$

(S1)

where $m_{ij}$ is the total amount of raw materials acquired from industry $j$ per unit of product in industry $i$, $m_{ijk}$ is the amount of raw material $k$ acquired from industry $j$ per unit of product in industry $i$, $v_{ij}$ the value of raw materials acquired from industry $j$ per unit of product in industry $i$, and $v_i/v_j$ is the unit value of products in industry $i/j$. Assuming the input-output efficiency is 100%, the material balance in industry $i$ is

$$\sum_j m_{ij} = 1 \qquad (S2)$$

3) Activity data of energy consumption

For a certain industry, the amount of energy consumption per unit of product is calculated based on the total energy consumption, the total production value, and the unit value of products, such that

$$e_{ix} = \frac{E_{ix}v_i}{V_i} \qquad (S3)$$

where $e_{ix}$ is the amount of energy $x$ consumption per unit of product in industry $i$, $E_{ix}$ is total energy $x$ consumption in industry $i$, and $V_i$ is total production value in industry $i$.

4) Activity data of transportation

For a certain industry, based on its province-wise/city-wise data of production value, the spatial distribution is determined. With the activity data of raw material, the probabilistic transportation distance for transferring one raw material to one province/city is sampled, which obeys the probability distribution

$$P(D_{im} = D_{nm}) = \frac{V_{in}}{V_i}$$

(S4)

where $D_{im}$ is the transportation distance for transferring raw material from industry $i$ to province/city $m$, $D_{nm}$ is the transportation distance from province/city $n$ to province/city $m$, and $V_{in}$ is the production value in industry $i$ in the province/city $n$.

5) Activity data of waste

For a certain industry, the amount of waste per unit of product is calculated based on the total amount of waste, the total production value, and the unit value of products, such that

$$w_{iy} = \frac{W_{iy}v_i}{V_i} \tag{S5}$$

where $w_{iy}$ is the amount of waste $y$ per unit of product in industry $i$, and $W_{iy}$ is total waste $y$ in the industry $i$.

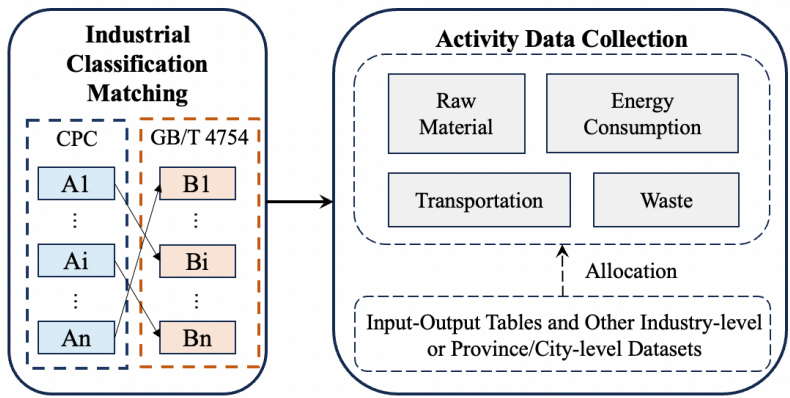

*SI Figure 1 | The flowchart of the activity data collection.*

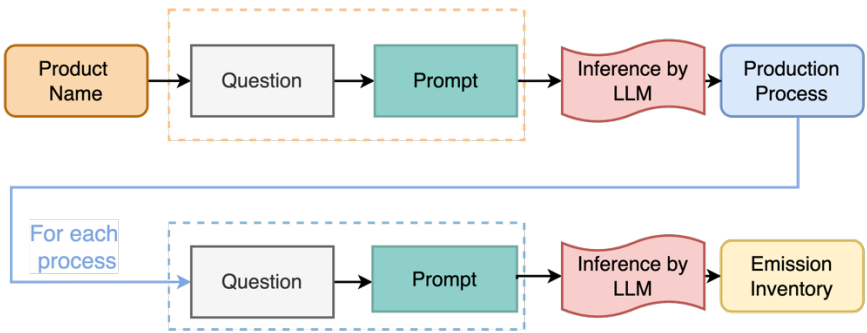

*SI Figure 2 | The flowchart of the production process and emission inventory generation*

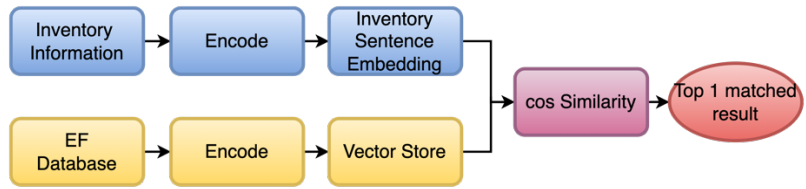

*SI Figure 3 | **The framework of the semantic-based emission factor matching model.***